\documentclass[10pt,twocolumn,letterpaper]{article}
\usepackage[pagenumbers]{cvpr}
\usepackage[T1]{fontenc}
\usepackage{graphicx}
\usepackage{amsmath,amssymb}
\usepackage{booktabs,multirow,array}
\usepackage{microtype}
\usepackage{longtable}
\usepackage{url}
\usepackage{seqsplit}
\definecolor{cvprblue}{rgb}{0.21,0.49,0.74}
\usepackage[pagebackref,breaklinks,colorlinks,allcolors=cvprblue]{hyperref}
\usepackage[capitalize]{cleveref}
\def\confName{CVPR}
\def\confYear{2027}

\newcommand{\method}{Select and Compensate}

\newcommand{\ours}{Ours}
\newcommand{\scorer}{\ensuremath{S_{\theta}}}
\newcommand{\ci}[2]{\ensuremath{[#1,\,#2]}}
\newcommand{\na}{\textemdash}
\newcommand{\supp}[1]{Supplement~#1}
\graphicspath{{figures/}}
\DeclareMathOperator{\LN}{LN}
\DeclareMathOperator{\GELU}{GELU}
\DeclareMathOperator{\vecop}{vec}
\DeclareMathOperator{\roundop}{round}

\makeatletter
\patchcmd{\@maketitle}{\null\iftoggle{cvprrebuttal}{\vspace*{-.3in}}{\vskip .375in}}{}{}{}
\patchcmd{\@maketitle}{\vspace*{24pt}}{\vspace*{12pt}}{}{}
\makeatother
\title{Carry the Fine Print:\\Budget-Aware Visual Token Compression for Document OCR}
\author{
  \small
  \begin{tabular}{@{}c@{\quad}c@{\quad}c@{}}
    {\large Yixuan Huang$^*$} & {\large ShengFan Zhang$^*$} & {\large Jiawei Chen$^\dagger$}\\
    University of Electronic Science & Beijing University of Posts & Central South University\\
    and Technology of China & and Telecommunications & \\
    {\footnotesize Chengdu, China\enspace\texttt{wangdaijiayou@163.com}} &
    {\footnotesize Beijing, China\enspace\texttt{15060126965@163.com}} &
    {\footnotesize Changsha, China\enspace\texttt{bleachranger@163.com}}\\[3pt]
    \multicolumn{3}{c}{\footnotesize $^*$ Equal contribution; $^\dagger$ Corresponding author}
  \end{tabular}
}
\hypersetup{pdftitle={Carry the Fine Print: Budget-Aware Visual Token Compression for Document OCR},pdfauthor={Yixuan Huang, ShengFan Zhang, Jiawei Chen}}
\begin{document}
\maketitle

\begin{abstract}
Visual token compression often treats salience or similarity as a proxy for transcription importance. Complete document OCR breaks this assumption: small symbols and repeated entries can be indispensable to the transcript despite being visually inconspicuous. We present Select and Compensate, a two-stage framework built on a simple idea: choosing where to retain a token need not determine all the evidence it carries. Selection combines adaptive encoding resolution with global importance and local contrast to choose which tokens to retain as anchors. Compensation enriches these anchors with nearby discarded content and its relative position, without adding tokens. The OCR backbone remains frozen, and native coordinates and structural tokens are preserved. We evaluate complete page transcription across five backbones on OmniDocBench and document reconstruction on MPDocBench. With half the original visual tokens, our input-adapted configuration reaches 95.42 OmniDocBench Overall on Ovis versus 96.09 for the uncompressed model, and paired evaluations show gains over simply reducing the input resolution on both benchmarks. With the original input grid, our method stays within 0.17 points of the uncompressed model at 75\% retention and exceeds a learned global selector by 4.70 points at 25\%. On a fixed 200-page Ovis workload at the same 50\% visual-token budget, only 3 of its outputs hit the length limit, versus 30--34 for the evaluated RTPrune, DivPrune, and HiPrune adaptations, and it finishes the batch 35--40\% sooner. Code will be made publicly available.
\end{abstract}

\section{Introduction}
\label{sec:intro}
End-to-end document OCR converts a whole page into prose, formulas, tables, and reading order. Unlike answering a question about one region, complete transcription requires evidence distributed throughout the page. High-resolution encoding exposes that evidence but produces long visual sequences that consume decoder context and cache. Compressing these sequences therefore raises a task-specific difficulty: what looks redundant in an image may still be essential to its transcript~\cite{informationhorizon2026,rtprune2026,layoutlite2026}.

\begin{figure}[t]
  \centering
  \includegraphics[width=\linewidth]{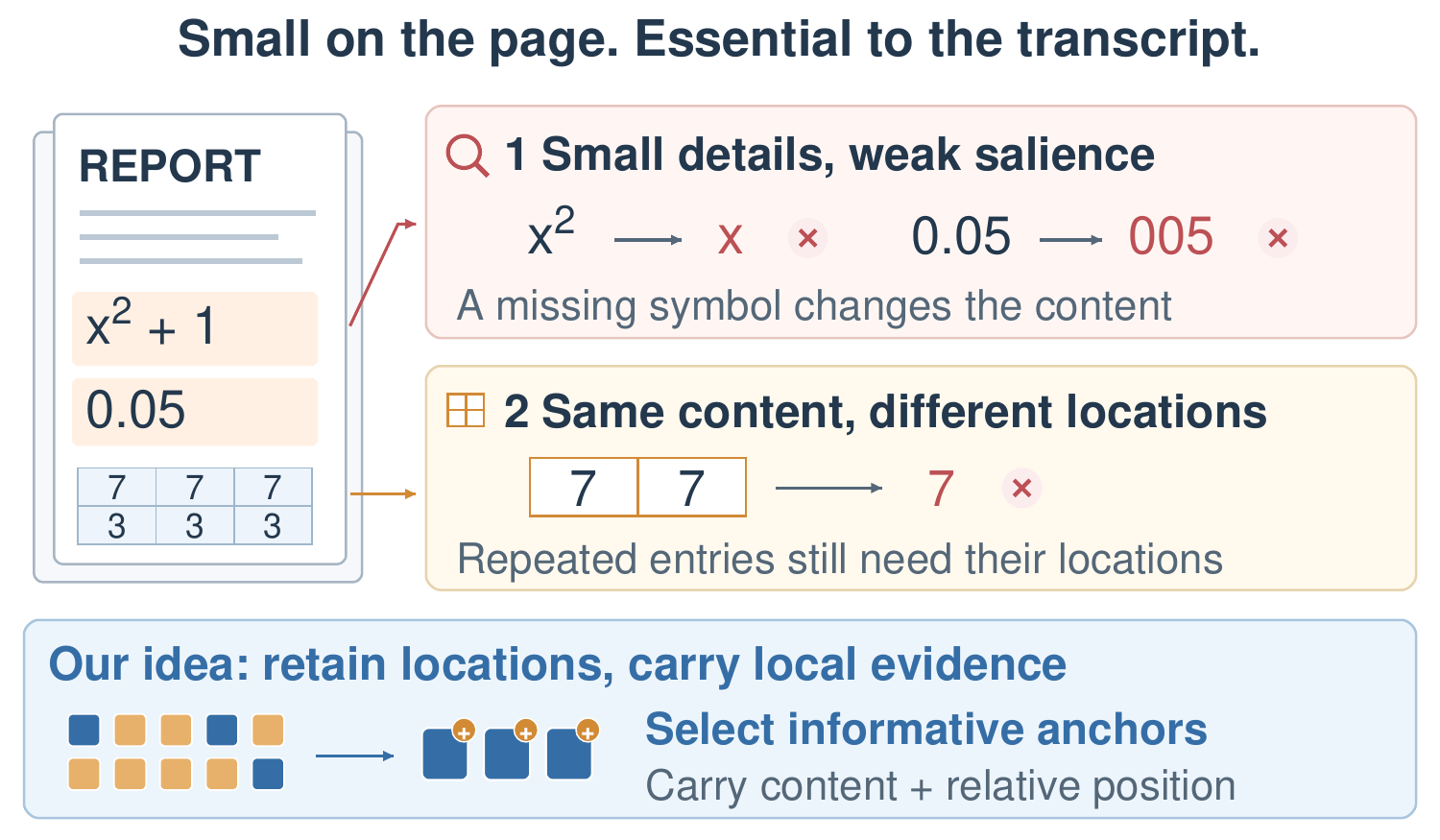}
  \caption{\textbf{Small on the page, essential to the transcript.} Schematic examples illustrate why small symbols and repeated entries cannot be judged by occupied area or appearance alone. Our design separates the locations retained by selection from the nearby content and relative-position evidence carried through compensation.}
  \label{fig:motivation}
\end{figure}

\paragraph{Challenge 1: small details can be easy to discard.}
A decimal point or superscript occupies little space but can change a number or formula. Global salience can favor a heading over such local distinctions, while uniform downsampling may remove them before token selection begins. The issue is not simply to retain diverse or important features~\cite{divprune2025,visionselector2026}, but to allocate a short sequence to evidence of very different spatial scales (\cref{fig:motivation}, top).

\paragraph{Challenge 2: fewer locations must represent dispersed evidence.}
Even a useful selection leaves encoded content outside its retained locations. Simply dropping it makes that evidence unavailable to the decoder. Aggregation offers a way to carry it forward~\cite{prumerge2025,rtprune2026}, but content alone does not describe its role: identical digits in different table cells are separate entries, and a symbol's position relative to nearby text can change its meaning (\cref{fig:motivation}, middle).

The field has made clear progress: learned selectors, layout-aware scoring, feature merging, and adaptive inputs show that visual sequences can shrink while retaining substantial capability~\cite{divprune2025,visionselector2026,prumerge2025,rtprune2026,layoutlite2026}. For complete OCR, however, visual salience remains coupled to transcription importance, input resolution to decoder budget, and a retained location to the evidence it carries. Page-level averages hide document structure and length-limited failures.

These gaps motivate two coupled questions: \emph{where should tokens remain explicit, and what evidence should each retained token carry?} We call each retained token an \emph{anchor} and distinguish its spatial \emph{address} from its carried \emph{evidence}. One decoder slot exposes one location, while its embedding can also encode nearby content and where that content lay. This lets us design location selection and local evidence aggregation together, rather than treating every removed location as discarded information.

\begin{figure*}[t]
  \centering
  \includegraphics[width=\textwidth]{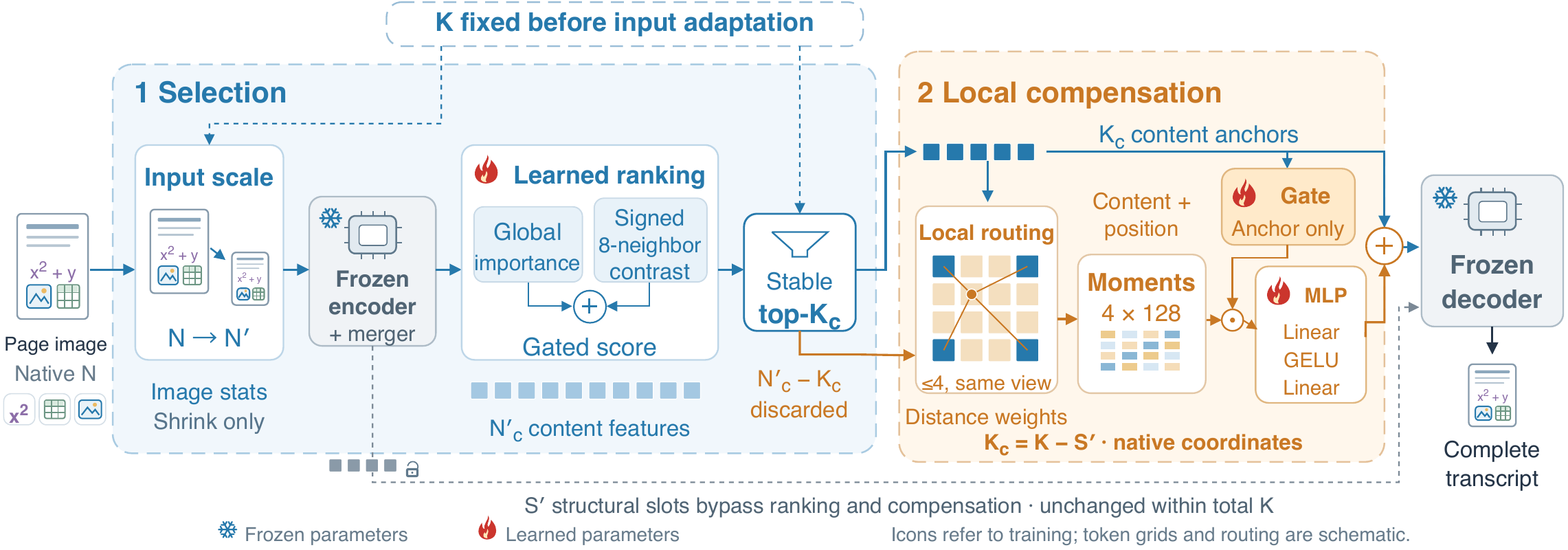}
  \caption{\textbf{Allocate addresses, carry nearby evidence.} Selection adjusts the input before the frozen encoder/merger, then ranks content features to retain $K_c=K-S'$ anchors. Compensation turns discarded content--position evidence into an anchor-gated residual. The $S'$ structural slots of the processed grid bypass ranking and compensation, giving exactly $K$ decoder slots, the original total budget. Native coordinates refer to the processed grid. A real-page input example and geometry are in \supp{S13}.}
  \label{fig:method}
\end{figure*}

\method\ implements this idea in two stages (\cref{fig:method}). \emph{Selection} fixes the decoder sequence length from the original page, adapts encoding resolution, and combines learned global importance with gated neighborhood contrast to retain informative anchors. \emph{Compensation} routes discarded features to nearby anchors and encodes their content and relative positions in an anchor-conditioned residual. The compact sequence preserves native coordinates and structural slots, so the backbone stays frozen.

We evaluate complete transcription on five OCR backbones and on multi-page document reconstruction. At half the original visual sequence, our input-adapted Ovis configuration reaches 95.42 Overall against 96.09 for the uncompressed model and improves over reduced resolution on both benchmarks with positive paired intervals. Compression also changes how transcripts end: only 15 of its 1,651 pages reach the output-length cap, versus 29--347 for the other compressed configurations.

Our contributions are: (i) a framework that couples spatial address allocation with local evidence aggregation for complete OCR; (ii) a realization through adaptive selection and local content--position compensation that keeps the OCR backbone frozen and adds no decoder slots; and (iii) evaluation on five OCR backbones and two benchmarks showing that, with original-grid inputs, the method comes within .17 points of the uncompressed model at 75\% retention and exceeds a learned global selector by 4.70 points at 25\%, 3.57 of them from compensation, and that on multi-page documents it raises cross-page table TEDS over that selector by 6.60 points on Logics.

\section{Related Work}
\label{sec:related}
\paragraph{Visual-token selection and aggregation.}
Visual-token compression has progressed from similarity merging and decoder-layer pruning~\cite{tome2023,fastv2024,sparsevlm2025,pyramiddrop2025} to pre-decoder selection by attention, learned importance, or diversity~\cite{visionzip2025,hiprune2025,visionselector2026,divprune2025}, and to learned objectives that unify pruning and merging~\cite{metacompress2026}. Methods that keep discarded evidence merge it into retained or summary tokens by feature similarity or optimal transport~\cite{prumerge2025,visionzip2025,rtprune2026}, or pool it into a variational residual trained jointly with LoRA~\cite{divico2026}. Our compensation instead pools discarded content locally, jointly with each discarded token's signed offset from its receiving anchor, so a retained token carries both nearby content and where that content lay; an anchor-only gate weights the pooled moments, and the frozen decoder receives no added tokens. For anchor allocation, we extend learned global importance~\cite{visionselector2026} with gated local contrast.

\paragraph{Document compression and OCR.}
RTPrune~\cite{rtprune2026} prunes DeepSeek-OCR~\cite{deepseekocr2025} tokens and merges the remainder through optimal transport, LayoutLite~\cite{layoutlite2026} trains a lightweight token scorer with layout supervision while preserving native coordinates, and DocKylin combines row/column removal with clustering~\cite{dockylin2025}. These methods move compression toward document structure, while end-to-end OCR systems require complete structured output~\cite{donut2022,nougat2023,pix2struct2023,gotocr2024}. Reduced resolution remains a strong alternative~\cite{vtcbench2026}. Our setting fixes the original decoder budget before input adaptation and protects structural slots.

\paragraph{Resolution, interfaces, and evaluation.}
Memory tokens, pooling/resampling, mixed-granularity pruning, and hard top-$K$ selection use different state budgets and interfaces~\cite{visco2026,qmop2026,bifocal2026,sheafprune2026}. Dynamic-resolution systems adapt image size or tiling~\cite{qwen2vl2024,internvl152024,llavauhd2024}, often changing the visual grid and decoder sequence together. Page and document benchmarks expose complementary structure, while format-decoupled OCR targets uncertainty in structured output~\cite{omnidocbench2024,mpdocbench2026,fdrl2026}. We use both at a fixed original budget.

\section{Select and Compensate}
\label{sec:method}
\method\ follows one representation principle: keep a small set of spatial addresses explicit, and let each address carry local evidence. Under the original total budget $K$, selection allocates addresses and compensation enriches them with discarded evidence. The two stages share this budget; compensation adds no tokens.
\subsection{Problem and native interface}
\label{sec:interface}
Let $N=N_c+S$ be the original total of content and structural visual slots; structural slots are positions the backbone inserts without page content. At retention $r$, the total decoder budget is
\begin{equation}
 K=\min\{N,\max\{1,\roundop(rN)\}\}.
 \label{eq:budget}
\end{equation}
Ties round to even. Input adaptation never recomputes $K$. The frozen encoder and merger then produce $N'=N_c'+S'$ native slots. Let $\mathcal C$ index its $N_c'$ content features $x_i\in\mathbb R^D$ with processed-grid coordinates $p_i=(h_i,w_i)$. All $S'$ structural features are copied unchanged; the content budget is
\begin{equation}
 K_c=K-S',\qquad |\mathcal A|=K_c,\quad \mathcal A\subseteq\mathcal C.
 \label{eq:contentbudget}
\end{equation}
Admissible inputs satisfy $S'<K\leq N'$. Only content features enter selection and compensation. Retained content and structural slots are restored to source order and keep their positions on the processed grid, and text positions continue as if the processed sequence were uncompressed. At full retention the sequence is unchanged.

A native view is one encoded image grid. Ovis, Logics, and Qwen use one view per page without structural slots. For an $H\times W$ merged grid, Hunyuan adds a newline slot after each row plus begin and end slots, so $N=H(W+1)+2$; a processed $H'\times W'$ grid gives $S'=H'+2$ and $K_c=K-H'-2$. Unlimited-OCR encodes a global view and, when tiled, a stitched local view, each with a newline slot per row, plus one separator (\supp{S17}).

\subsection{Budget-aware selection: input resolution}
\label{sec:resolution}
For single grids without structural slots, selection counts low-variation cells on the original native RGB canvas. Each cell is the $32\times32$ (Ovis/Logics) or $28\times28$ (Qwen) pixel region of one merged token and is flat when its 8-bit grayscale intensity range is at most 12 and its population standard deviation at most 3. Flatness is a visual proxy for blank background. With $F$ flat cells, the target encoded area is
\begin{equation}
 N^*=\begin{cases}
 N, & N-F\leq K,\\
 NK/(N-F), & N-F>K.
 \end{cases}
 \label{eq:resolution}
\end{equation}
The first branch keeps the native grid; the second reduces area toward fitting non-flat evidence within $K$. Candidate grids $h\times w$ do not exceed the original dimensions, satisfy $K\leq hw\leq N$, and have log aspect-ratio deviation at most .05. We pick the grid whose $hw$ is closest to $N^*$, breaking ties by smaller aspect deviation, larger $hw$, smaller height, and smaller width, and resize the original RGB once with bicubic interpolation.

Hunyuan reserves each candidate grid's $H'+2$ structural slots within $K$. Unlimited-OCR measures its actual padded views in 64-pixel cells and searches uniformly rescaled inputs through the unchanged official tiler; admissibility uses total $N'$ and $S'$. These adapter-specific rules preserve the same original budget (\supp{S13, S17}).

\subsection{Budget-aware selection: learned ranking}
\label{sec:selector}
The ranking scorer \scorer\ extends the learned global importance scorer (LIS) of VisionSelector~\cite{visionselector2026}, with affine projections $q_i=Q(x_i)$ and $k_i=K_p(x_i)$ to width $d=512$. Its global score is
\begin{equation}
 s_i=\frac{1}{N_c'\sqrt d}\sum_{j\in\mathcal C}q_i^\top k_j,\qquad i\in\mathcal C.
 \label{eq:lis}
\end{equation}
Equivalently, $s_i=q_i^\top\bar k/\sqrt d$, where $\bar k$ averages content keys. Global compatibility supplies page context; local contrast exposes distinctions from nearby features. Let $\mathcal N(i)\subset\mathcal C$ contain existing eight-connected neighbors in the same native view, excluding $i$. Define
\begin{equation}
 u_i=k_i-\frac{1}{|\mathcal N(i)|}\sum_{j\in\mathcal N(i)}k_j,
 \qquad c_i=\frac{q_i^\top u_i}{\sqrt d}.
 \label{eq:contrast}
\end{equation}
Boundary tokens use their actual neighbor count; an empty neighborhood has $u_i=0$. The gated score is
\begin{equation}
 g_i=\sigma\!\left(w_g^\top[q_i;u_i]+b_g\right),
 \qquad s'_i=s_i+\alpha g_i c_i.
 \label{eq:ranking}
\end{equation}
Distinct query/key projections make the signed effect content-dependent. Zero-initialized $\alpha,w_g,b_g$ add $2d+2$ parameters and preserve the inherited global ranking initially. Only $\alpha$ initially receives a gradient through this branch; the gate can learn once $\alpha$ changes.

Stable descending sorting selects the $K_c$ largest content scores, breaking ties by source index and restoring source order. Write $D_0=\mathcal C\setminus\mathcal A$ for discarded content. The $S'$ structural slots bypass this operation.

\subsection{Local content--position compensation}
\label{sec:compensation}
Selecting $K_c$ content anchors fixes their addresses. Compensation pools discarded evidence into a fixed-size residual at each anchor, preserving its coordinate on the processed grid.
For each native view $v$, let $\mathcal A_v$ be its anchors. Each discarded token (donor) $j\in D_0$ routes to the nearest $\min(4,|\mathcal A_v|)$ anchors in its view by squared native-grid distance, with stable tie breaking. Views without anchors have no receivers (\supp{S17}). Let $\mathcal R(j)$ denote the receivers. Coordinate-only routing uses
\begin{equation}
 \omega_{ji}=\frac{\|p_j-p_i\|_2^{-2}}
 {\sum_{a\in\mathcal R(j)}\|p_j-p_a\|_2^{-2}},\qquad i\in\mathcal R(j).
 \label{eq:route}
\end{equation}
Weights are zero outside $\mathcal R(j)$. Donors and anchors within a view have distinct coordinates, so all selected distances are positive. A donor can influence several anchors; the support is overlapping rather than a partition.

We project donor features to width 128 with $v_j=W_v\LN(x_j)$. Relative offsets define a bounded basis:
\begin{align}
 \delta^h_{ji}&=\tanh(h_j-h_i),\quad
 \delta^w_{ji}=\tanh(w_j-w_i),\nonumber\\
 \phi_{ji}&=(1,\delta^h_{ji},\delta^w_{ji},\delta^h_{ji}\delta^w_{ji}).
 \label{eq:basis}
\end{align}
For each anchor, four vector moments jointly pool content and relative position:
\begin{align}
 \mu_i&=\sum_{j\in D_0}\omega_{ji},\nonumber\\
 m_{i\ell}&=\frac{\sum_{j\in D_0}\omega_{ji}\phi_{ji\ell}v_j}
 {\sqrt{1+\mu_i}},\quad \ell=0,\ldots,3.
 \label{eq:moments}
\end{align}
The zeroth moment carries pooled content; the remaining moments couple it to signed local offsets. These fixed-size statistics add evidence without adding decoder slots. The anchor alone determines the gate $G_i=\operatorname{reshape}(\sigma(W_g\LN(x_i)))\in\mathbb R^{4\times128}$; gated moments then pass through a learned residual map:
\begin{equation}
 \widetilde x_i=x_i+
 W_o\GELU\!\left(W_h\vecop(G_i\odot m_i)\right),\quad i\in\mathcal A.
 \label{eq:residual}
\end{equation}
The hidden width is 256. Zero-initialized $W_o$ makes compensation initially preserve the selected embeddings. The decoder receives $(\widetilde x_i)_{i\in\mathcal A}$ at the original anchor positions, with any structural slots inserted unchanged.

\subsection{Training and implementation}
\label{sec:training}
The visual encoder, merger, and language decoder remain frozen. Selector training uses the inherited cardinality-constrained soft mask and transcript cross-entropy, together with the LIS soft-to-hard mask constraint. The soft mask keeps all $N$ visual slots during selector training; exactly $K$ are hard-selected at inference. After selector training, we freeze its parameters and train compensation on hard selected anchors using transcript cross-entropy.

The evaluated recipe inherits a backbone-specific global scorer, adds 2,000 \scorer\ updates, then fits compensation for 2,000 updates. The selector cycles through $.25/.50/.75$ retention, and compensation is fitted at each evaluated rate, with backbone-specific schedules in \supp{S2}. Each backbone has its own targets and weights; new stages use seed 42 and the final checkpoint. We train at native resolution and adapt the encoding input only at inference.

The global scorer materializes an $N_c'\times N_c'$ QK matrix, neighborhood lookup costs $O(N_c'\log N_c'+N_c'd)$, and compensation compares donors with all anchors, sorting distances in chunks of 256; \cref{sec:cost} measures the overhead.

\section{Experiments}
\label{sec:experiments}
We organize the evidence around four questions: (1) how much complete-page quality the representation retains at the main 50\% budget and across 25--75\% retention; (2) whether the same design extends to document structure and multiple OCR interfaces; (3) how selection and compensation contribute; and (4) how visual-budget reduction changes output tails and measured request cost. Cross-task transfer is reported separately in \supp{S11}.
\begin{table*}[t]
\centering
\caption{\textbf{Complete-method comparison on OvisOCR2, 1,651 pages, original 50\% decoder budget.} \ours\ is the complete method with input adaptation; baselines keep their own preprocessing and runtime. The uncompressed model and \ours\ keep the state of the decoder's state-space (SSM) layers in FP32, the other methods in BF16. LayoutLite matches mean-page retention, and reduced resolution approximates the budget with a smaller dense grid. Visual tokens are decoder slots relative to the original $\sum N$; capped pages reach the output-length limit and remain scored. Bold/underline: best/second-best compressed result in each column.}
\label{tab:main}
\small\setlength{\tabcolsep}{5pt}
\begin{tabular}{lrrrrrrr}
\toprule
Method & Visual tokens (\%) & Overall $\uparrow$ & Text edit $\downarrow$ & CDM $\uparrow$ & TEDS $\uparrow$ & Order edit $\downarrow$ & Capped pages $\downarrow$\\
\midrule
Uncompressed & 100.0 & 96.085 & 0.0251 & 95.972 & 94.797 & 0.1105 & 10\\
\midrule
RTPrune & 50.0 & 73.132 & 0.4359 & 80.235 & 82.747 & 0.2776 & 347\\
DivPrune & 50.0 & 77.219 & 0.2192 & 78.324 & 75.255 & 0.2091 & 186\\
HiPrune & 50.0 & 90.900 & 0.1142 & 94.426 & 89.696 & 0.1665 & 134\\
LayoutLite & 52.1 & 92.500 & 0.0748 & 94.261 & 90.721 & 0.1557 & 115\\
LIS & 50.0 & 94.073 & 0.0552 & \underline{94.746} & 92.992 & 0.1375 & 65\\
Reduced resolution & 50.0 & \underline{94.821} & \underline{0.0336} & 94.185 & \underline{93.637} & \underline{0.1183} & \underline{29}\\
\midrule
\ours\ & 50.0 & \textbf{95.417} & \textbf{0.0270} & \textbf{94.892} & \textbf{94.057} & \textbf{0.1148} & \textbf{15}\\
\bottomrule
\end{tabular}

\end{table*}

\subsection{Evaluation protocol}
\label{sec:protocol}
\paragraph{Data and supervision.}
As our setting requires, all backbones are single-stage: one model reads the whole page image, with no layout detector or crop recognizer. OvisOCR2~\cite{ovisocr2model}, Logics-Parsing-V3~\cite{logicsparsingv3model}, and HunyuanOCR-1.5~\cite{hunyuanocrmodel} use all 1,651 OmniDocBench v1.6 pages. Qwen2.5-VL~\cite{qwen25vl2025} and Unlimited-OCR~\cite{unlimitedocrmodel} share a frozen 384-page subset, Omni384: four mutually exclusive 96-page strata (tables, formulas, reading order, and remaining text), selected from annotation metadata with source/language coverage and seed 42 before Unlimited-OCR evaluation. All Qwen arms run through one common Hugging Face implementation (Qwen-HF). Outputs, scores, failures, and timing do not determine page selection, and all arms within a backbone use the same pages. MPDocBench (MP) supplies 409 available documents/3,051 pages; 11 documents/84 pages are unavailable. Compression training uses 4,343 olmOCR-mix-1025~\cite{olmocr2report2025} pages, split into 3,875 train and 468 calibration pages with disjoint source families. OCR-specialized models learn from their own uncompressed transcripts; Qwen uses the dataset's released transcriptions (\texttt{natural\_text}). Evaluation labels never train compressors, and an audit found no confirmed duplicates between training and evaluation pages (\supp{S1}).

\paragraph{Comparisons and metrics.}
The primary budget is $K/N=.50$ of the original $N$. We compare with RTPrune~\cite{rtprune2026}, DivPrune~\cite{divprune2025}, HiPrune~\cite{hiprune2025}, LayoutLite~\cite{layoutlite2026}, LIS (our OCR-trained adaptation of VisionSelector's global scorer~\cite{visionselector2026}), and reduced resolution; each keeps its own input processing and runtime (\supp{S2}). Original-grid configurations skip input adaptation and keep the backbone's default preprocessing. Exact-$K$ methods share the original decoder budget; LayoutLite and reduced resolution approximate it (\cref{tab:main}). Omni Overall averages text quality $100(1-\mathrm{edit})$ from normalized edit distance, formula CDM~\cite{cdm2025}, and table TEDS~\cite{teds2020}; reading-order edit is separate. Text, formula, table, and order are scored on 1,557/313/458/1,638 of the 1,651 pages. Every page, length ending, repeat, and parsing failure remains included.

\paragraph{Uncertainty.}
Paired bootstrap uses 1,000 seed-42 draws over 1,443 source groups for the 1,651 pages, 382 for Omni384, or 409 MP documents. Intervals are marginal 95\% unless called adjusted; adjusted ones keep their Bonferroni level (98.75\% for input-adapted Ovis, 97.5\% for input-adapted MP, 99.17\% for Unlimited-OCR). Differences without an interval are point estimates.

\subsection{Complete OCR at a fixed decoder budget}
\label{sec:ocr}

\paragraph{What changes under compression?}
\Cref{tab:main} shows uneven quality loss at the same original $K$. RTPrune and DivPrune lose text, formula, table, and order quality, with 347/186 length-limited outputs versus 10 for the uncompressed model and 15 for \ours, the fewest among compressed configurations. HiPrune and LayoutLite narrow the quality gap but still leave 134 and 115 pages at the cap, and learned LIS retains substantially more quality. Saved failures and fixed positive/negative cases connect these results to complete transcription (\supp{S6, S9}).

\paragraph{Complete-recipe quality.}
The complete method reaches 95.417 Overall (\cref{tab:main}), retaining most of the uncompressed 96.085 with half the visual slots, and is the best compressed configuration on every component metric. It exceeds the measured RTPrune, DivPrune, HiPrune, and LayoutLite adaptations by 22.285, 18.198, 4.516, and 2.917 points, and reduced resolution by .596 points, adjusted interval \ci{.067}{1.159}. Relative to the uncompressed model, text is nearly lossless (edit .0270 versus .0251) and reading order stays close (.1148 versus .1105); the remaining loss lies mainly in formulas and tables, at CDM 94.892 versus 95.972 and TEDS 94.057 versus 94.797. With original-grid inputs, its advantages over LIS, HiPrune, DivPrune, and LayoutLite all have positive paired intervals (\supp{S4, S6}); the input-adapted interval family is in \supp{S12}.

\paragraph{Quality across compression strengths.}
\Cref{tab:retention} reports the original-grid configuration at all three retention rates on the full 1,651 pages. The advantage over LIS grows as compression strengthens, from .493 points at 50\%, marginal 95\% interval \ci{.133}{.830}, to 4.703 points at 25\%, interval \ci{3.709}{5.725}, where \ours\ also exceeds RTPrune by 50.982 points, interval \ci{49.034}{53.013}. Every component favors \ours\ over LIS at this budget: formula CDM 89.062 versus 83.119, table TEDS 86.850 versus 84.480, and text/reading-order edit .1357/.1942 versus .1936/.2256. The largest per-page gain at 25\% comes from a Putnam problem page, where LIS runs to the 16,384-token cap and \ours\ stops after 2,391 tokens (\supp{S8}). At 75\%, \ours\ reaches 95.918, only .167 points below the uncompressed model, and exceeds RTPrune by 2.406 points, interval \ci{2.080}{2.774}; at this light compression it is on par with LIS. Its formula CDM, 96.148, even exceeds the uncompressed 95.972, text and reading-order edit stay at .0266 and .1126 versus .0251 and .1105, and the gap lies mainly in tables (TEDS 94.266 versus 94.797). Reduced resolution, which changes the encoded grid instead of selecting tokens, reaches 90.389 at 25\%. Full controls are in \supp{S4}.

\begin{table}[t]
\centering\small
\caption{\textbf{Ovis quality at 25\%, 50\%, and 75\% retention, all 1,651 pages.} Overall on original-grid inputs. The \ours\ scorer is fixed across rates, with compensation trained per rate. Compressed arms keep BF16 SSM state, unlike the input-adapted FP32 configuration in \cref{tab:main}. Bold/underline: best/second-best token-selection method in each row.}
\label{tab:retention}
\begin{tabular}{crrr}
\toprule
Retention & LIS & RTPrune & \ours\\
\midrule
25\% & \underline{82.745} & 36.466 & \textbf{87.448}\\
50\% & \underline{94.073} & 73.132 & \textbf{94.566}\\
75\% & \underline{95.887} & 93.512 & \textbf{95.918}\\
\midrule
100\% & \multicolumn{3}{c}{96.085 (uncompressed)}\\
\bottomrule
\end{tabular}

\end{table}

\paragraph{Does the framework extend to other OCR interfaces?}
\Cref{tab:transfer} compares complete configurations with independently adapted compression methods on five backbones spanning single grids, structural row slots, and tiled views. \ours\ has the best compressed score on Ovis, Logics, and Unlimited-OCR and is second to LIS on Hunyuan, and it retains 95.1--99.3\% of the uncompressed Overall on four of the five backbones. On HunyuanOCR, it exceeds DivPrune, RTPrune, and HiPrune by 4.400, 15.818, and .403 points while preserving the original structural-row budget; with original-grid inputs, its gains over RTPrune and DivPrune have positive intervals \ci{14.063}{17.460} and \ci{3.200}{5.417}. On Logics, it exceeds RTPrune, DivPrune, and HiPrune by 28.063, 24.196, and 4.360 points. Unlimited-OCR adds a tiled multi-view interface: \ours\ reaches 89.892 on Omni384 versus 77.149, 44.335, and 51.983 for RTPrune, DivPrune, and HiPrune, with 9 length-capped outputs versus 39, 100, and 55, and its original-grid gains over RTPrune and DivPrune, 12.564 and 45.378 points, have positive adjusted intervals \ci{8.187}{17.386} and \ci{39.631}{51.379} (\supp{S17}). On Qwen, \ours\ exceeds RTPrune, DivPrune, and LIS but trails HiPrune and LayoutLite (\cref{sec:discussion}). Component scores and the VisCo adaptation are in \supp{S14, S17}.

\begin{table*}[t]
\centering
\caption{\textbf{Complete-method comparison across five backbones, original retention $.50$.} \ours\ includes input adaptation; baselines keep their own preprocessing and runtime. Qwen-HF runs all arms through one common Hugging Face implementation; Unlimited-OCR and Qwen-HF use Omni384. Rel.\ is \ours\ as a percentage of the same backbone's uncompressed Overall. LayoutLite, which requires layout training for each backbone, is compared on Ovis in \cref{tab:main}; the VisCo adaptation is in \supp{S14}. Bold/underline: best/second-best compressed score in each row.}
\label{tab:transfer}
\footnotesize\setlength{\tabcolsep}{4pt}
\begin{tabular}{lllrrrrrrr}
\toprule
Backbone & Visual interface & Pages & Uncompressed & RTPrune & DivPrune & HiPrune & LIS & \ours & Rel.\ (\%)\\
\midrule
Ovis & Single grid & 1,651 & 96.085 & 73.132 & 77.219 & 90.900 & \underline{94.073} & \textbf{95.417} & 99.3\\
Logics-V3 & Single grid & 1,651 & 94.729 & 64.645 & 68.511 & 88.348 & \underline{91.516} & \textbf{92.708} & 97.9\\
HunyuanOCR & Grid + row slots & 1,651 & 95.092 & 75.252 & 86.671 & 90.667 & \textbf{91.654} & \underline{91.070} & 95.8\\
Unlimited-OCR & Global + tiles & 384 & 94.483 & 77.149 & 44.335 & 51.983 & \underline{89.409} & \textbf{89.892} & 95.1\\
Qwen-HF & Single grid & 384 & 76.184 & 58.414 & 61.045 & \textbf{68.556} & 58.925 & \underline{62.285} & 81.8\\
\bottomrule
\end{tabular}

\end{table*}

\begin{table}[t]
\centering\small
\caption{\textbf{Matched-input configurations, original $.50$ budget.} Global scorer: the inherited global scorer without learned contrast or compensation. Both columns share processed pixels, actual $K$, and runtime; $\Delta$ is \ours$-$Global in points.}
\label{tab:strongcontrol}
\begin{tabular}{lrrr}
\toprule
Backbone & \ours & Global scorer & $\Delta$\\
\midrule
Ovis & 95.417 & 95.524 & $-0.107$\\
Logics-V3 & 92.708 & 92.981 & $-0.273$\\
HunyuanOCR & 91.070 & 91.312 & $-0.242$\\
Unlimited-OCR (384) & 89.892 & 89.556 & $+0.336$\\
Qwen-HF (384) & 62.285 & 65.136 & $-2.851$\\
\bottomrule
\end{tabular}

\end{table}

\begin{table}[t]
\centering\small\setlength{\tabcolsep}{3pt}
\caption{\textbf{Configuration ablations on Ovis, original 50\% retention.} Comp.: local compensation. Comp.\ only applies the same frozen compensation to fixed equally spaced anchors at the original $K$. Ranking only uses the learned ranking without input adaptation or compensation, with BF16 SSM state versus FP32 for the complete method.}
\label{tab:adopted}
\label{tab:controls}
\begin{tabular}{llcr}
\toprule
Configuration & Selection & Comp. & Overall\\
\midrule
Complete & Budget-aware & $\checkmark$ & 95.417\\
Comp.\ only & Uniform anchors & $\checkmark$ & 91.663\\
Ranking only & Learned only & \na & 94.672\\
\bottomrule
\end{tabular}

\end{table}

\subsection{Document-level evaluation}
\label{sec:documents}
Document evaluation uses 409 MPDocBench documents/3,051 pages and all eight official columns, including Figure; Ovis assembles single-page requests, while Logics uses its official two-page requests with carried state. With original-grid inputs, \ours\ improves document Overall over LIS by .594 points on Ovis, interval \ci{.251}{.914}, and by 2.072 points on Logics, interval \ci{1.329}{2.930}, and over RTPrune by 18.391 and 30.479 points (\cref{tab:mpversions}). Its text, table TEDS, and reading-order differences against LIS also have positive intervals on both backbones.

\begin{table}[t]
\centering\small
\caption{\textbf{Document quality on MPDocBench, 409 documents.} Overall averages all eight official columns; X-text and X-TEDS are the cross-page text and table columns. Original-grid rows follow each backbone's request format; both input-adapted Ovis arms keep FP32 SSM state. Bold/underline: best/second-best within each backbone and block. All eight columns are in \supp{S7, S14}.}
\label{tab:mpversions}
\setlength{\tabcolsep}{4pt}
\begin{tabular}{llrrr}
\toprule
Backbone & Method & Overall & X-text & X-TEDS\\
\midrule
\multicolumn{5}{l}{\emph{Original-grid input}}\\
Ovis & LIS & \underline{67.522} & \underline{71.704} & \underline{64.616}\\
Ovis & RTPrune & 49.725 & 40.126 & 56.510\\
Ovis & \ours & \textbf{68.116} & \textbf{72.484} & \textbf{64.663}\\
Logics & LIS & \underline{70.188} & \underline{80.256} & \underline{77.195}\\
Logics & RTPrune & 41.782 & 36.775 & 58.787\\
Logics & \ours & \textbf{72.261} & \textbf{83.106} & \textbf{83.799}\\
\midrule
\multicolumn{5}{l}{\emph{Input-adapted}}\\
Ovis & Reduced resolution & 68.514 & 74.896 & 64.414\\
Ovis & \ours & \textbf{68.872} & \textbf{75.240} & \textbf{64.698}\\
\bottomrule
\end{tabular}

\end{table}

The cross-page columns are the most document-specific. On Logics, \ours\ improves cross-page table TEDS over LIS by 6.604 points, interval \ci{3.399}{10.035}, and cross-page text by 2.850 points, interval \ci{1.189}{4.782}. Cross-page table TEDS is also the largest single contribution to Logics' 2.072-point document Overall gain over LIS, adding .826 points, followed by text and cross-page text at .371 and .356. The gain is therefore visible in linking text and tables across page boundaries, not only in a page average. With input adaptation in both arms, Ovis reaches 68.872 Overall and exceeds reduced resolution by .358 points, adjusted 97.5\% interval \ci{.070}{.668}; all eight columns and their supports are in \supp{S7, S14}.

\subsection{Stage effects and gain sources}
\label{sec:controls}
\paragraph{Intervening on the trained stages.}
In the Ovis configuration ablation (\cref{tab:controls}), the complete method at 95.417 Overall exceeds compensation on fixed equally spaced anchors (91.663) by 3.754 points and learned ranking alone (94.672) by .745 points, with the original per-page $K$ on all 1,651 pages. On the same adapted input, it also exceeds rule-based selection without learned ranking or compensation (90.909) by 4.508 points, adjusted interval \ci{3.395}{5.688} (\supp{S12}). The large drop with uniform anchors indicates that anchor placement still matters when discarded evidence is carried (\supp{S14}).

\paragraph{When does carried evidence matter?}
Compensation pays off where retained addresses are scarce or weak. With original-grid input and the learned ranking fixed, adding compensation at 25\% retention raises Ovis from 83.880 to 87.448, a 3.568-point gain with interval \ci{2.530}{4.481} that accounts for most of the 4.703-point advantage over LIS in \cref{tab:retention}. All four components improve: text edit falls from .181 to .136 and reading-order edit from .221 to .194, while formula CDM rises from 84.772 to 89.062 and table TEDS from 85.002 to 86.850. On Qwen, removing it under the same input lowers Overall from 62.285 to 55.503, with text edit rising from .212 to .324, formula CDM falling from 64.192 to 57.721, and table TEDS from 43.863 to 41.236. Where addresses are plentiful, its effect is small. At 50\%, removing compensation changes Ovis and Logics by less than .1 point, and a global scorer without learned contrast or compensation that shares each complete method's processed input, $K$, and runtime stays within .3 points on Ovis, Logics, and Hunyuan (\cref{tab:strongcontrol}); it is lower on Unlimited-OCR and higher on Qwen. An added Overall advantage over this control is therefore not established at 50\%. At this budget, the input allocation that both share matters more: removing only the input-adaptation step from the complete method lowers Ovis by .832 points at matched FP32 runtime (\supp{S4--S5, S12, S14}).

\begin{table*}[t]
\centering\small\setlength{\tabcolsep}{6pt}
\caption{\textbf{Ovis costs at a 50\% visual budget, original-grid inputs, 200 pages.} Omni Overall is the quality of the same original-grid configurations on all 1,651 OmniDocBench pages (\cref{tab:main,tab:retention}); the 200 timing pages are not quality-scored. TTFT: time to first token; E2E: end-to-end request latency; times in seconds. Capped: length endings in batch-1 runs. Burst: all 200 requests submitted at once, with request-completion p50 and whole-batch wall time. Used KV and GPU: sampled burst peaks of used key--value (KV) cache and of total allocated memory, which is dominated by the preallocated cache. Compressed arms keep BF16 SSM state, the uncompressed model FP32. One RTX 4080 SUPER. Bold/underline: best/second-best among compressed methods; ties share a mark.}
\label{tab:cost}
\begin{tabular}{lrrrrrrrrrr}
\toprule
 & Visual & Omni & TTFT & \multicolumn{2}{c}{E2E} & Capped & \multicolumn{2}{c}{Burst} & Used KV & GPU\\
\cmidrule(lr){5-6}\cmidrule(lr){8-9}
Method & tokens (\%) & Overall $\uparrow$ & p50 & p50 & p95 & pages & p50 & wall & (MiB) & (MiB)\\
\midrule
Uncompressed & 100 & 96.085 & .147 & 2.94 & 8.40 & 0 & 52.9 & 71 & 9,788 & 26,758\\
\midrule
LIS & 50 & \underline{94.073} & \textbf{.157} & 3.03 & \textbf{8.86} & \textbf{1} & \textbf{43.2} & \textbf{121} & \textbf{5,229} & \underline{27,940}\\
RTPrune & 50 & 73.132 & .171 & \underline{2.63} & 74.83 & 30 & 43.3 & 191 & 6,282 & 27,944\\
DivPrune & 50 & 77.219 & .238 & \textbf{2.43} & 74.85 & 31 & 54.7 & 206 & 6,491 & 27,950\\
HiPrune & 50 & 90.900 & .216 & 2.90 & 75.18 & 34 & 53.3 & 204 & 6,484 & \textbf{27,474}\\
\ours & 50 & \textbf{94.566} & \textbf{.157} & 3.01 & \underline{8.93} & \underline{3} & \textbf{43.2} & \underline{124} & \textbf{5,229} & 27,946\\
\bottomrule
\end{tabular}

\end{table*}

\paragraph{Inference budgets and sensitivity.}
A frozen $.50$ checkpoint applied at $.25/.50/.75$ retention to 128 development pages reaches 89.380/94.815/96.614; three predeclared flatness thresholds span .350 points at $.50$ (\supp{S16}).

\paragraph{Where does input adaptation help?}
Input adaptation improves the complete Ovis/Logics/Qwen configurations over their original-grid counterparts by .851/.717/2.658 points; Hunyuan's improves by .132 under a separately validated serial, unchunked-prefill runtime, and Ovis also changes its SSM precision. Partitioning pages by selection's image-and-budget decision, without using outputs or ground truth and with the original metric denominators, places almost all of these gains on resized pages: .820 of Ovis's .851 points come from 714 resized pages and .031 from 937 unchanged grids, and resized pages contribute .673 of Logics' .717 and 2.902 of Qwen's 2.658. This matches the rule's intent: keep resolution where shrinking is unnecessary and relieve content--budget pressure elsewhere (\supp{S13, S15}).

\subsection{Measured costs and output tails}
\label{sec:cost}
\paragraph{Budget and completion benefits.}
With original-grid inputs, \ours\ halves the visual sequence and cuts the sampled peak of used key--value (KV) cache by 46.6\% relative to the uncompressed model, from 9,788 to 5,229\,MiB (\cref{tab:cost}), and its median burst request completes earlier, at 43.2 versus 52.9\,s. Against RTPrune, DivPrune, and HiPrune at the same budget, it has a .157\,s median TTFT versus .171--.238\,s, an 8.93\,s E2E p95 versus about 75\,s, and a 17--19\% lower used-KV peak, and it finishes the burst in 124\,s instead of 191--206\,s, 35--40\% sooner. Burst wall time tracks the length-capped requests of the same run: with 0, 1, and 3 of them, the uncompressed model, LIS, and \ours\ finish in 71, 121, and 124\,s, whereas RTPrune, DivPrune, and HiPrune, with 29--30 each, need 191--206\,s. The same holds per request: in batch-1 runs, the three capped pages of \ours\ account for 189 of the 200\,s by which its cumulative page latency exceeds the uncompressed model's, and the other 197 pages add 10.8\,s. A few runaway 16,384-token generations thus decide when a batch completes, so avoiding them saves request time as well as transcript quality.

\paragraph{Quality at matched cost.}
With its full-benchmark quality column, \cref{tab:cost} separates the methods on both axes. LIS and \ours\ share the original-grid input, visual budget, and runtime and differ only in the learned ranking and compensation. At the same median TTFT, median burst completion, and used-KV peak, these add .493 points on all 1,651 pages, with a positive interval (\cref{tab:retention}); RTPrune, DivPrune, and HiPrune are 3.666--21.434 points lower and need 191--206\,s rather than 124\,s to finish the batch.

\paragraph{Operator and training overhead.}
In a separate component pass that decodes one token per page, the complete compression path of \ours\ takes 6.19\,ms median per page (ranking 1.99\,ms, compensation 3.50\,ms), versus 20.5 and 82.9\,ms for the RTPrune and DivPrune operators; its 200-page total of 1.24\,s is about 0.14\% of the 909\,s cumulative page latency. Both stages run once per page before decoding, so their cost does not grow with output length. Input adaptation adds a median 101.8\,ms per page for image statistics and resizing (\supp{S12}). Fitting the inherited global scorer, 2,000 ranking updates, and 2,000 compensation updates on Ovis takes about 2.65 GPU-hours, excluding generation of the uncompressed training transcripts (\supp{S2, S8}).

\section{Conclusion}
\label{sec:discussion}
Select and Compensate treats retained positions as addresses whose embeddings can carry nearby evidence. With half the visual sequence, the input-adapted Ovis configuration reaches 95.417 Overall against 96.085 uncompressed and exceeds reduced resolution on pages and documents with positive adjusted intervals. The three original-grid retention rates show how the two stages divide the work. At 75\%, selection nearly suffices: even LIS stays within .2 points of the uncompressed model, and \ours\ within .17. As the budget tightens, carried evidence takes over and supplies 3.57 of the 4.70-point advantage over LIS at 25\%. Across page boundaries, the gain is largest on Logics' cross-page tables. On HunyuanOCR's row slots and Unlimited-OCR's tiled views, \ours\ exceeds RTPrune, DivPrune, and HiPrune. Across settings, fewer runaway outputs accompany higher quality and earlier batch completion. Complete transcription must therefore be judged by quality and completion together: a compact representation must specify which locations remain addressable and how discarded evidence reaches them, not only how many tokens remain.

\paragraph{Limitations.}
Gains depend on backbone: on Qwen, \ours\ has the best text edit, .212 versus .367/.272 for HiPrune/LayoutLite, but lower formula and table scores (\supp{S14}); results on dots.ocr~\cite{dotsocrmodel} are in \supp{S6, S14}. The input rule was developed after inspecting OmniDocBench and MPDocBench results, so its generalization to unseen sources remains untested. Main training uses one seed, and the OCR-trained compression weights do not preserve QA/VQA performance without QA training (\supp{S11}).

Linking anchor selection with local evidence aggregation, \method\ provides a compact representation for complete page and document transcription.

\clearpage
\section*{LLM Usage Statement}
OpenAI Codex with GPT models (including GPT-6.1-Sol) and Anthropic Claude were used as coding and writing agents under the authors' direction. They wrote most of the code, including the implementation and the data-preparation, experiment-orchestration, and analysis scripts; ran experiments and diagnosed failures; drafted and revised much of the manuscript, supplement, and tables; checked literature and citations; and produced figures. GPT-6-Astra and Claude also gave reviewer-style feedback on the manuscript. The research ideas, method, and experimental design came primarily from the authors, who chose the datasets and made all scientific decisions; benchmark annotations were not generated by these tools. The authors reviewed and verified all tool-generated content and take full responsibility for the submission.
{
\small
\bibliographystyle{ieeenat_fullname}
\bibliography{references}
}
\end{document}